\makeatletter\def\graphicscache@inhibit{true}\makeatother

\documentclass[letterpaper, 10 pt, conference]{ieeeconf}  %

\IEEEoverridecommandlockouts                              %

\usepackage{siunitx}

\usepackage[hidelinks]{hyperref}
\usepackage[style=ieee,hyperref,natbib=true,backend=bibtex,firstinits,doi=false,%
     mincitenames=1,maxcitenames=2,maxbibnames=99,sorting=none,terseinits=false,hyperref=true]{biblatex}
\bibliography{references.bib}
\renewbibmacro*{bbx:savehash}{}%
\defbibheading{bibliography}[\bibname]{\section*{References}}

\usepackage{xstring}
\usepackage{graphicx}
\usepackage[export]{adjustbox}

\usepackage{tikz}
\usetikzlibrary{positioning}
\usetikzlibrary{shapes.geometric}
\usetikzlibrary{arrows.meta}
\usetikzlibrary{backgrounds}
\usetikzlibrary{calc}

\usepackage{pgfplots}
\usepgfplotslibrary{statistics}
\pgfplotsset{compat=1.14,
  /pgfplots/ybar legend/.style={
    /pgfplots/legend image code/.code={%
       \draw[##1,/tikz/.cd,yshift=-0.25em]
        (0cm,0cm) rectangle (3pt,0.8em);},
   },
}

\usepackage{booktabs}
\usepackage{threeparttable}

\IfFileExists{graphicscache.sty}{\usepackage{graphicscache}}

\usepackage{amsmath}
\usepackage{tensor}

\newcommand{\T}[2]{\ensuremath{\tensor[^{#1}]{T}{_{#2}}}}
\newcommand{\TT}[2]{\ensuremath{\tensor[^{#1}]{\tilde{T}}{_{#2}}}}

\newcommand{\R}[2]{\ensuremath{\tensor[^{#1}]{R}{_{#2}}}}
\newcommand{\RT}[2]{\ensuremath{\tensor[^{#1}]{\tilde{R}}{_{#2}}}}

\usepackage{placeins}

\usepackage[hidelinks]{hyperref}
\usepackage[capitalize]{cleveref}

\usepackage[draft]{fixme}
\fxsetup{theme=color, layout={inline,index}}

\usepackage[firstpage=true]{background}
\newcommand\copyrighttext{%
\parbox{\textwidth}{
\footnotesize
\textbf{Accepted final version.} IEEE International Symposium on Safety, Security, and Rescue Robotics (SSRR), Abu Dhabi,
UAE, to appear Nov 2020
}
}
\SetBgContents{\copyrighttext}
\SetBgScale{1}
\SetBgColor{black}
\SetBgAngle{0}
\SetBgOpacity{1}
\SetBgPosition{current page.north}
\SetBgVshift{-0.8cm}
\usepackage[bottom]{footmisc}

\title{Autonomous Wall Building\\with a UGV-UAV Team at MBZIRC 2020}
\author{Christian Lenz$^{*}$, Max Schwarz$^{*}$, Andre Rochow, Jan Razlaw,\\ Arul Selvam Periyasamy, Michael Schreiber, and Sven Behnke%
\thanks{\hspace{-2.2ex}$^{*}$: equal contribution.}%
\thanks{This work has been supported by a grant of the Mohamed Bin Zayed International Robotics Challenge (MBZIRC).}%
\thanks{Institute for Computer Science VI, Autonomous Intelligent Systems, University of Bonn, Endenicher Allee 19a, 53115 Bonn, Germany,
		{\tt\small \{lenz, schwarz\}@ais.uni-bonn.de}%
}
}

\begin{document}

\maketitle              %
\thispagestyle{empty}
\pagestyle{empty}

\begin{abstract}
Constructing large structures with robots is a challenging task with many potential applications that requires mobile manipulation capabilities.
We present two systems for autonomous wall building that we developed for the
Mohamed Bin Zayed International Robotics Challenge 2020.
Both systems autonomously perceive their environment, find bricks, and build
a predefined wall structure.
While the UGV uses a 3D LiDAR-based perception system which measures
brick poses with high precision, the UAV employs a real-time camera-based
system for visual servoing.
We report results and insights from our successful participation at the
MBZIRC 2020 Finals, additional lab experiments, and discuss the lessons learned
from the competition.
\end{abstract}

\section{Introduction}

Mobile manipulation is needed to handle objects in large work spaces, e.g. for constructing structures.
While ground-based mobile manipulation has received considerable research attention,
robotic aerial manipulation is still in its infancy.
The Mohamed Bin Zayed International Robotics Challenge (MBZIRC)~2020\footnote{\url{http://mbzirc.com/}}
posed tasks for robot teams in a demanding outdoor setting.
In its Challenge~2, participants were required to build walls out
of supplied bricks, both with a UGV and a team of up to three UAVs.
The task setting was particularly interesting, because it required complete
autonomy, robustness under real-world outdoor conditions with harsh sunlight
and wind, and independence from any outside reference system besides the
globally available GPS.

\begin{figure}
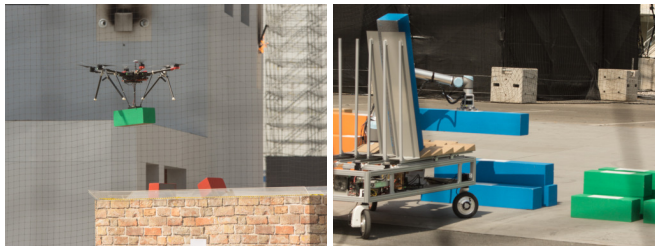

 \centering
 \includegraphics[clip,trim=400 0 200 300,  height=3.2cm]{images/finn/MBZIRC_2020-137.jpg}%
 \hfill%
 \includegraphics[clip,trim=410 250 390 200,height=3.2cm]{images/finn/MBZIRC_2020-65.jpg}\vspace*{-1ex}
 \caption{Our UAV Lofty (left) and UGV Bob (right) during the MBZIRC 2020 Finals.}
 \label{fig:lofty_bob}
\end{figure}

In this work, we describe our entry to the MBZIRC 2020 Finals, which consists
of a UGV-UAV team (see \cref{fig:lofty_bob}). In addition to describing our integrated systems for solving
the tasks set by the competition and discussing lessons learned, our technical
contributions include:

\begin{itemize}
 \item a flexible and precise magnetic gripper system for large objects
    addressing the unique design constraints on UAVs,
 \item a robust and efficient vision-based detection and pose estimation module
    for box-shaped objects,
 \item a laser-based pose estimation and registration module for
    the UGV, and
 \item a highly space- and time-efficient box storage system for UGVs.
\end{itemize}

\section{MBZIRC 2020}

\begin{figure}[t]
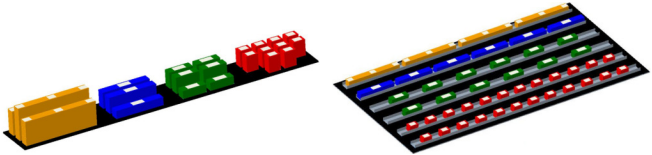

 \centering
 \includegraphics[clip, width=.49\linewidth]{images/rules/bob_pile.jpg}
 \hfill%
 \includegraphics[clip, width=.49\linewidth]{images/rules/lofty_pile.jpg}%
 \vspace*{-2ex}
 \caption{Brick pickup arrangement for the UGV (left) and UAVs (right).}
 \label{fig:challenge2}
\end{figure}

In Challenge~2 of the MBZRIC 2020 competition, a team of one UGV and
up to three UAVs had to pick, transport, and place bricks to build a wall.
Four different brick types with 20$\times$20\,cm cross-section were used:
Red (30\,cm length, 1\,kg), green (60\,cm, 1.5\,kg), blue (120\,cm, 1.5\,kg),
and orange (180\,cm, 2\,kg). Each type of robot had a designated pick-up and
place area inside the arena (40$\times$50\,m). \cref{fig:challenge2} shows the
arrangement of the bricks for the UGV and UAVs at the beginning of the task.
Both robots had to build the first wall segment using only orange bricks. For the
remaining segments (one for the UGV and three for the UAVs), a random blueprint defining the order of the red, green,
and blue bricks was provided some minutes before the competition. Points were granted for correctly placed bricks. The UGV could archive
between 1 to 4 points per brick (45 bricks in total); the bricks placed by
an UAV counted between 3 to 16 points (46 bricks). The time limit for this
challenge was 25\,min. All tasks had to be performed autonomously to archive
the perfect score. The teams were allowed to call a reset at
any time to bring the robots back to the starting location. Resets did not result in
a point penalty, but no extra time was granted.

\section{Related Work}

\textit{UGVs for Wall Building:}
The application of robots for wall-building has a long history~\citep{slocum1988blockbot}.
One particularly impressive example is the work of Dorfler et al.~\cite{dorfler2016mobile}, who developed a heavy mobile bricklaying robot for the creation of free-form curved walls.
An alternative for creating free-form walls is on-site 3D printing with a large manipulator arm~\cite{keating2017toward}.

\textit{UGVs for Disaster Response:}
Our work mostly relates to disaster
response robotics, where protective or otherwise functional structures have
to be built quickly and with minimal human intervention.
The DARPA Robotics Challenge~\citep{krotkov2017darpa} established a baseline
for flexible disaster response robots.
In comparison to these and to more recent disaster-response robots such as Centauro~\cite{Klamt:RAM19}, our system has a much higher degree of autonomy,
but is much more specialized for the task at hand.

\textit{Aerial Manipulation:}
In recent years, aerial manipulation has become a research focus~\Citep{ruggiero2018aerial}.
Complex systems with fully actuated multi-DoF robotic arms have been built~\citep{huber2013first,kim2013aerial}.
Lindsey et al.~\cite{lindsey2012construction} demonstrated the assembly of structures with teams of small UAVs. This work relied on an external motion capture system and self-locking magnetic part connectors.
Goessens et al.~\cite{goessens2018feasibility} present a feasibility study of constructing real-scale structures with UAVs, which is based on self-aligning Lego-like brick shapes.

A predecessor of our work is Challenge~3 of the last MBZIRC edition in
2017, where a team of UAVs was supposed to collect discs.
Our entry \citep{beul2019team} was quite successful and reached a third place
in this challenge, behind ETH Zurich~\citep{bahnemann2019eth} and CTU Prague, UPENN and UoL~\citep{spurny2019cooperative}.
In comparison, the 2020 edition of MBZIRC featured much heavier and larger objects,
which could only be grasped on a specific spot and had to be placed
in a specified pose.
To this end, we designed a magnetic gripper that is guided using visual servoing
and has five passive DoFs that allow flexibility during grasping but facilitate
rigid and precise placement.

\section{UGV Solution}

We build our ground robot Bob based on our very successful UGV which won
the first MBZIRC competition~\citep{schwarz2019team}. We improved the
basis and adapted the manipulator and sensors for the new challenge.
Since 45 bricks had to be picked, transported, and placed
in 25\,min to obtain a perfect score,
we developed our UGV to store as many bricks as
possible. We decided to use a 3D LiDAR as the main sensor to detect and localize the
piles of bricks and partially built wall.

\subsection{Hardware Design}

\begin{figure}
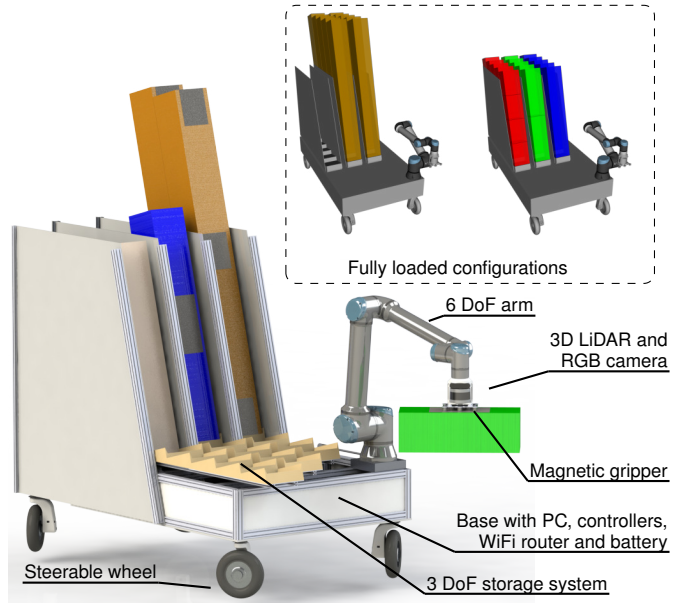

	\centering
﻿\begin{tikzpicture}[
 	font=\sffamily\scriptsize,
    every node/.append style={text depth=.2ex},
	box/.style={rectangle, inner sep=0.5, anchor=west},
	line/.style={black, thick}
]

\node[anchor=south west,inner sep=0] (image) at (0,0) {\includegraphics[height=7cm]{images/bob/Bob.png}};
\node[anchor=south west,inner sep=0] (image) at (3.8,4.9) {\includegraphics[height=3cm]{images/bob/storage_orange.png}};
\node[anchor=south west,inner sep=0] (image) at (6.3,4.9) {\includegraphics[height=3cm]{images/bob/storage_color.png}};

\node[box, align=center](rviz) at(4.5,4.5){Fully loaded configurations};
\draw[rounded corners, dashed] (3.7,4.3) rectangle (8.6,8.0);

\node[box, align=right, anchor=east](laser_scanner) at(8.8,3.4){3D LiDAR and\\RGB camera};
\draw[line](laser_scanner.south west)--(laser_scanner.south east);
\draw[line](laser_scanner.south west)--(6.4,2.9);

\node[box](arm) at(5.8,4.0){6 DoF arm};
\draw[line](arm.south west)--(arm.south east);
\draw[line](arm.south west)--(5.6,3.8);

\node[box](wheel) at(0.2,0.45){Steerable wheel};
\draw[line](wheel.south west)--(wheel.south east);
\draw[line](wheel.south east)--(2.7,0.25);

\node[box, align=right,, anchor=east](base) at(8.8,1.0){Base with PC, controllers,\\WiFi router and battery};
\draw[line](base.south west)--(base.south east);
\draw[line](base.south west)--(4.4,1.45);

\node[box, align=right, anchor=east](gripper) at (8.8,1.8){Magnetic gripper};
\draw[line](gripper.south west)--(gripper.south east);
\draw[line](gripper.south west)--(6.2,2.6);

\node[box, align = left,  anchor=east](storage) at (8.0,0.3){3 DoF storage system};
\draw[line](storage.south west)--(storage.south east);
\draw[line](storage.south west)--(3.1,1.9);

\end{tikzpicture}

 	\vspace*{-2ex}
	\caption{UGV hardware design. Top right: The storage system is capable of holding either 10 orange bricks (left)
	or all remaining bricks (20 red, 10 green, 5 blue) (right).}
	\label{fig:bob}
\end{figure}

UGV components include a  four-wheeled omnidirectional
base, a 6-DoF manipulator arm with custom-made magnetic gripper, a wrist sensor consisting of a 3D LiDAR as well as an RGB camera, and a 3-DoF storage system.

The base has a footprint of 1.9$\times$1.4\,m to provide enough space for our storage
system. It rolls on direct-drive brushless DC hub motors, controlled by two ODrive driver boards.
Since the motors were originally intended for hover boards, i.e. personal conveyance devices,
they have enough torque to accelerate the approx. 90\,kg UGV. To achieve
omnidirectional movement, we coupled each wheel with a Dynamixel H54-200-S500-R servo which
rotates it around the vertical axis.
The developed base supports driving speeds of up to 4\,m/s and precise positioning for
manipulation tasks.

This year's competition required to manipulate objects of up to 1.80\,m length
and to stack them to a total height of 1\,m. Instead of the UR5 mounted on our
2017 robot, we used a UR10e 6-DoF robotic
arm, which gives us the benefit of a larger workspace (1.30\,m) and
enough payload capability (10\,kg) to carry the gripper, including the sensors, and the
bricks (up to 2\,kg). We adapted the arm controller to work with UGV
battery power.

Bob's gripper is equipped with eight electromagnet and a contact switch.
Since the ferromagnetic parts of the bricks are very thin (approx. 0.6\,mm), we
decided to use a larger number of smaller magnets to distribute the contact surface as much as
possible while keeping the total gripper size minimal.
The switch detects if a brick is securely grasped.

For perceiving the bricks, we mounted a Velodyne VLP-16 3D LiDAR
and a Logitech Brio camera on the wrist. The LiDAR is our main sensor for
detecting the bricks and for estimating their poses relative to the robot (see
\cref{sec:brick_detection}). The RGB images are used to detect the wall
marker (see \cref{sec:wall_marker}).
A second Logitech Brio camera mounted at the top of the robot provides the
operator situation awareness.

We designed a storage system
which has three individually actuated storage compartments. Each compartment has five
bins to store bricks. The ground plate of each bin is 20,5\,cm wide, 20\,cm
long and is mounted inclined 15$^\circ$ backwards. This inclination forces the
bricks to slide in a known pose inside the storage system even if the bricks are grasped
imprecise. Hence, we do not need an additional perception system to perceive
the current pose of the stored bricks.
Side walls hold the bricks in place during UGV movements.
The walls are 110\,cm high which is sufficient to
hold the largest bricks (180\,cm long) in place. Furthermore, this system allows to
stack multiple small bricks (up to 4 red bricks, and up to 2 green bricks) to
increase the number of bricks to be stored in the system. Overall, the system is
capable to store either all large bricks (10 orange), or all remaining bricks
(20 red, 10 green, 5 blue, see \cref{fig:bob}).
Since the storage system exceeds the workspace of the
UR10e, each compartment can be moved horizontally (using a Dynamixel Pro L42-10-S300-R and a linear belt drive)
to put the desired bin in reach of the arm.

The UGV is equipped with a standard ATX mainboard with a quad-core
Intel Core i7-6700 CPU and 64\,GB RAM. The whole system is powered by an
eight-cell LiPo battery with 20\,Ah and 29.6\,V nominal voltage. This allows the
robot to operate for roughly one hour, depending on the task.

Due to resource
conflicts when building all robots needed for the MBZIRC 2020 competition, hardware and software
component development and testing was initially executed on the modified Mario robot~\citep{schwarz2019team}.
The larger Bob chassis was assembled on site in Abu Dhabi for the first time.

\subsection{High-level Control}

We implemented a high-level controller consisting of a finite-state machine (FSM)
generating the robot actions, a database to keep track of
every brick relevant for the UGV, and an algorithm computing the time-optimal
strategy for a given build order.

The FSM includes 32 different states for locomotion, manipulation,
perception, storage logistics, and fallback mechanisms.
After executing an action, the resulting database and FSM state
was stored to enable quick recovery after a reset.

Since the UR10 arm is very precise, we can manipulate multiple bricks from a stationary
position after perceiving the environment just once. Our overall strategy was
to minimize locomotion between different positions in front
of the piles and the wall. Thus, the plan was to pick up all orange bricks and bring them
to the wall at once. After successfully building the orange wall segment, the UGV
was to collect all remaining bricks to build the second wall segment.
Whereas the orange wall segment (two stacks of 5 bricks each) can be built from
two predefined positions, the build order of the second wall segment highly
depends on the supplied blueprint and can be optimized to minimize the number
of locomotion actions.

We implemented a backtracking algorithm to find the optimum build order. To make
this approach feasible regarding runtime, we only consider building the wall from
left to right, but allow starting the next layer before finishing the first.
Let the longer wall axis (from left to right) be denoted as the x-axis. First,
we compute the set of possible place positions by $P = \{x_i + t_x | x_i = \text{center of brick $i$}\}$.
The place pose is shifted by the arm reach $t_x = 0.675\text{\,m}$ to place the
robot such that the number of brick placement poses in reach is maximized.
Due to the wall structure, we have $ 7 \leq |P| \leq 35 $. We now enumerate all possible
ordered sequences $S \subseteq P$. For each $p_i \in S$, we build all bricks which meet the following
criteria:
\begin{enumerate}
 \item The brick was not built already,
 \item the brick is in reach based on the position $p_i$,
 \item the brick is fully supported by the ground or previously built bricks, and
 \item the left adjacent brick was built.
\end{enumerate}

\noindent $S = (p_1, p_2, \dots)$ is a valid solution if all bricks are built. We search
for the optimal solution with $|S|$ and $d_S$ minimal,
where $d_S = \sum_{i=2}^{|S|} |p_i - p_{i-1}|$, i.e. the shortest path to
traverse between all building positions.
Pruning sub-trees is used to accelerate the algorithm.

\subsection{Brick and Wall Perception}
\label{sec:brick_detection}

\begin{figure}[t]
	\centering
﻿\scalebox{0.9}{
\begin{tikzpicture}[font=\sffamily\scriptsize,on grid,>={stealth[inset=0pt,length=4pt,angle'=45]}]
\tikzset{every node/.append style={node distance=3.0cm}}
\tikzset{terminal_node/.append style={minimum size=1.5em,minimum height=3em,minimum width={width("Search Point")+0.2em},draw,align=center,rounded corners}}
\tikzset{content_node/.append style={minimum size=1.5em,minimum height=3em,minimum width={width("Search Point")+0.2em},draw,align=center,fill=blue!15!white, rounded corners}}
\tikzset{label_node/.append style={near start}}
\tikzset{group_node/.append style={align=center,rounded corners,inner sep=1em,thick}}
\tikzset{decision_node/.append style={align=center,shape aspect=1.5,minimum width=7.9em,minimum height=5.4em,diamond,draw,fill=yellow!25!white,font=\sffamily\normalsize,node distance=3.9cm}}

\definecolor{red}{rgb}     {0.5,0.0,0.0}
\definecolor{green}{rgb}   {0.0,0.5,0.0}
\definecolor{blue}{rgb}    {0.0,0.0,0.5}
\definecolor{grey}{rgb}    {0.5,0.5,0.5}

\node(marker_detection)[content_node] at (0.0, 0.0) {Pile/Marker\\Detection};
\node(render)[content_node] at (3.0,0.0){Render and\\Sample};
\node(rough_allign)[content_node] at (6.0, 0.0){Rough\\Alignment};
\node(rerender)[content_node] at(6.0, -1.8){Render and\\Sample Bricks};
\node(multi_align)[content_node] at (3.0, -1.8){Brick\\Alignment};
\node(conf)[content_node] at (0.0, -1.8){Confidence\\Estimation};

\draw[->, thick] (-2, 0.0) -- node[label_node,midway,above] {Trigger} (marker_detection);
\draw[->, thick] (marker_detection) -- node[label_node,midway,above] {$\T{B}{W,P}$} (render);
\draw[->, thick] (render) -- node[label_node,midway,above] {$PC_m$}(rough_allign);
\draw[->, thick] (rough_allign) -- node[label_node,midway,left] {$\TT{B}{W,P}$} (rerender);
\draw[->, thick] (rerender) -- (multi_align);
\draw[->, thick] (multi_align) -- node[label_node,midway,above] {$\TT{B}{b_i}$}(conf);
\draw[->, thick] (conf) -- node[label_node,midway,above]{Result} (-2,-1.8);

\end{tikzpicture}
}
 	\caption{LiDAR-based brick perception pipeline.}
	\label{fig:lidar_brick_perception}
\end{figure}
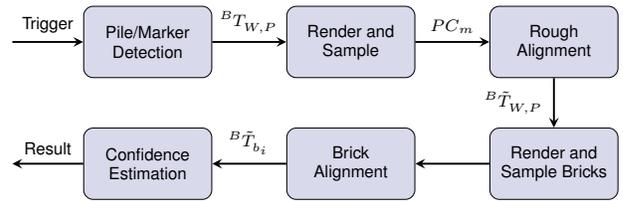

When the robot is close to either the pick-up location (called \textit{pile})
or the place location (called \textit{wall}), it needs to localize against
these objects and to perform pose estimation of the individual bricks in
order to pick them or place new bricks next to them.

Our perception pipeline assumes knowledge of the current state of the world,
including a rough idea of the brick poses relative to the pile or wall.
The perception pipeline receives this information from the high-level control module.

Depending on the target (pile/wall), the perception pipeline receives an initial
guess of the target pose $\T{B}{P}$ or $\T{B}{W}$ w.r.t. the robot's base ($B$).
It also receives the brick pose $\T{W,P}{b_i}$ and brick type $t_i \in \{\textrm{r, g, b, o}\}$
for each brick $i$.
For initial alignment purposes, the individual brick alignment can be switched off. Finally, bricks can be excluded from the optimization, for example
if they are far away and not of interest for the next action.

\Cref{fig:lidar_brick_perception} shows the overall perception process.
In both cases, it starts with a rough detection of the target location from
further away.

\subsubsection{Rough Pile Detection}

\begin{figure}
	\centering
	\includegraphics[clip,height=3.8cm,trim=40 80 80 70,frame]{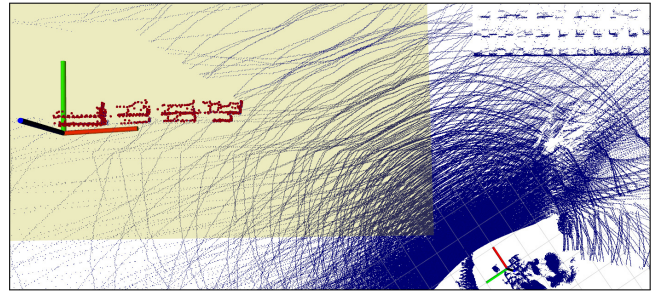}%
	\caption{Top down view for pile detection from LiDAR points (blue).
	The robot is located at the small coordinate system (bottom).
	The search area can
	be restricted using geofencing (yellow rectangle).
	Detected points are visualized in red and the estimated pile pose is shown.
	}
	\label{fig:pile_detection}
\end{figure}

In case of the pile, we know the approximate shape beforehand. We make use
of this and search for corresponding measurements using the 3D LiDAR sensor.
While doing so, one needs to take care to disambiguate the UGV and UAV piles
and other distractors in the arena.
The first step in detecting the pile is to confine the search space to a
user-defined search area, so-called geofencing.
We start by filtering out the points that lie outside of the search area and fit a plane to the remaining points. Next, we filter out the points that lie on the plane or are very close to the plane. The remaining points, shown in \cref{fig:pile_detection}, may belong to the pile.
After clustering the points and filtering clusters which do not fit the
expected pile size, we perform PCA on the remaining cluster to estimate the
largest principal component and define the pile coordinate system such that the
$X$ axis is aligned with the 2D-projected principal axis and the $Z$ axis points vertically upwards.

\subsubsection{Marker Detection}
\label{sec:wall_marker}
\begin{figure}
	\centering
	\includegraphics[clip,height=3.2cm]{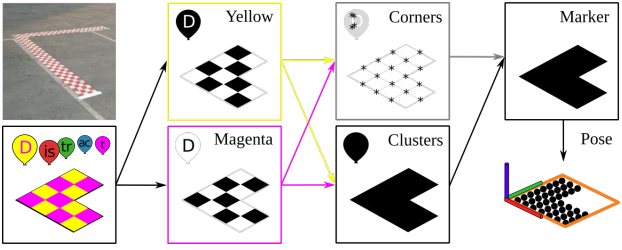}\vspace*{-1ex}
	\caption{Wall marker detection.
	Starting from the input image (Col.~1), two color masks are generated (Col.~2). These masks are used for extracting corners (Col.~3 top) and clustering (Col.~3 bottom). Corners vote for clusters to detect the wall marker (Col.~4 top). Marker pose is estimated using oriented bounding box (in orange around projected points col.~4 bottom).}
	\label{fig:wall_marker_detection}
\end{figure}

After picking up bricks, the next task is finding and estimating the pose of the L-shaped marker indicating where to build the wall (see \cref{fig:wall_marker_detection}).
Our idea for detecting the marker relies on its distinctive color, pattern and shape best visible in camera images.
We start by specifying volumes within the HSV color space corresponding to the yellow and magenta tones of the marker.
Now, we exploit the characteristic color composition of the marker to filter out distractors in the image.
For that, we generate a color mask using all yellow pixels that are close to magenta pixels and another one for magenta pixels in the vicinity of yellow pixels (\cref{fig:wall_marker_detection} Col.~2).
The resulting masks preserve the pattern of the marker which we utilize to filter out further distractors.
First, we extract the corners from each mask separately and then search for corners present in close vicinity in both masks.
Additionally, we fuse both masks and extract clusters of all masking pixels (\cref{fig:wall_marker_detection} Col.~3).
Next, we let each resulting corner vote for its corresponding cluster.
The cluster gathering most votes is assumed to be corresponding to the wall marker (\cref{fig:wall_marker_detection} Col.~4 top).

We project each cluster pixel onto the ground plane of the arena and
accumulate the resulting point clouds of the previous 10 seconds, since the camera has
limited FoV and we can make use of the robot and arm movements to cover more space.
After Euclidean clustering, we compute the smallest oriented 2D rectangle around
the biggest cluster. The intersection point of the L shape can be found by
looking for the opposite corner, which should have the highest distance from
all cluster points (see \cref{fig:wall_marker_detection} Col.~4 bottom).
Finally, the detection is validated by verifying the measured side lengths.

\subsubsection{Rendering and Sampling}

The next module in the brick perception pipeline (\cref{fig:lidar_brick_perception}) converts our parametrized world model into 3D point clouds that are suitable for point-to-point registration with the measurements $PC_s$ of the Velodyne 3D LiDAR, which is moved to capture a dense 3D scan of the pile or brick scene.
We render the parametrized world model using an OpenGL-based renderer~\citep{schwarz2020stillleben}
and obtain the point cloud $PC_m$.
Both point clouds are represented in the base-link $B$.
Since we render at a high resolution of 2800$\times$2800 pixels, we downsample the
resulting point cloud to uniform density using a voxel grid filter with resolution $d$ = 0.02\,m.

\subsubsection{Rough Alignment}
We will now obtain a better estimate $\TT{B}{W}$ or $\TT{B}{P}$ of the pile/wall pose.
We first preprocess $PC_s$ as follows:
\begin{enumerate}
	\item Extract a cubic region around $\T{B}{W}$/$\T{B}{P}$,
	\item downsample to uniform density of using a voxel grid filter with resolution 0.02\,m,
	\item find and remove the ground plane using RANSAC, and
	\item estimate point normals (flipped s.t. they point towards the scanner) from local
	   neighborhoods for later usage.
\end{enumerate}
\noindent We then perform Iterative Closest Point (ICP) with a
point-to-plane cost function \citep{low2004linear} with high correspondence distance,
which usually results in a good rough alignment, followed by a point-to-point
alignment with smaller correspondence distance for close alignment.

In case the wall marker was detected, we add another cost term
\vspace{-1ex}
\begin{equation}
 E_{dir}(\TT{B}{W}) = (1-(\RT{B}{W} \cdot (1\,0\,0)^T)^T \vec{l})^2
\end{equation}
with $\vec{l}$ being the front-line direction and $\RT{B}{W}$ the rotation component of $\TT{B}{W}$.
This cost term ensures the optimized wall coordinate system is aligned with the marker direction.

The above-defined cost function is optimized using the Ceres solver until either
the translation and rotation changes or the cost value  change are below
termination thresholds ($\lambda_{T} = \num{5e-8}$, $\lambda_{C} = \num{1e-6}$).

\subsubsection{Individual Brick Pose Optimization}
\label{sec:individual_bricks}

When the robot is close enough, we can determine individual brick poses.
We constrain the following optimization to translation and yaw angle (around the
vertical $Z$ axis), since pitch and roll rotations can only happen due to
malfunctions such as dropping bricks accidentally. In these cases, the brick
will most likely not be graspable using our gripper, so we can ignore
these cases and filter them later.

For correspondence information, we re-render the scene using the
pose $\TT{B}{W,P}$ obtained from rough alignment.
Here, we include the ground plane in the rendering, since we can use it to
constrain the lowest layer of bricks.
We separate the resulting
point cloud into individual brick clouds $PC_{bj}$.

We now minimize the objective
{\small\begin{equation}
E_{\text{multi}} = \sum_{j=1}^{N}\sum_{i=1}^{M(j)}
\frac{1}{M(j)} \|(R(\theta_j) p_{j,i} + t_j - q_{j,i})^T n_{q_{j,i}}\|^2
,
\end{equation}}
where the optimized parameters $\theta_i$ and $t_i$ describe the yaw angle and translation of brick $i$,
$N$ is the number of bricks, $M(j)$ is the number of found point-to-point
correspondences for brick $j$, $p_{j,i} \in PC_{bj}$ \& $q_{j,i} \in PC_{s}$
are corresponding points, and $n_q$ is the normal in point $q$.
This is a point-to-plane ICP objective with separate correspondences for each brick.
Correspondences are filtered using thresholds $\lambda_{dot}$ and $\lambda_{dist}$
for normal dot products and maximum point distances.

\begin{figure}[tb]
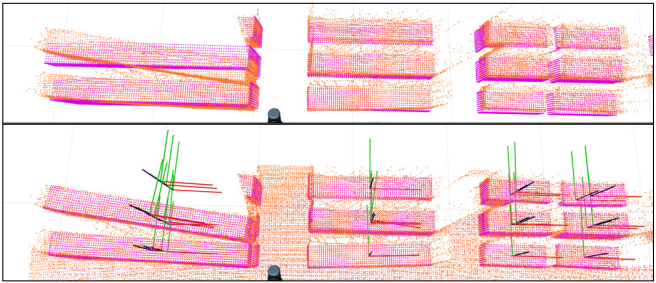

	\centering
	\includegraphics[width=\linewidth,clip,trim=0 100 0 70,frame]{images/perception/pile_above_rough.jpg}\\
	\includegraphics[width=\linewidth,clip,trim=0 100 0 0,frame]{images/perception/pile_above_multi.jpg}
	\caption{Precise alignment of individual bricks. Laser measurements are
	colored orange, model points are shown in purple.
	Top: Initial solution found by the rough ICP stage.
	Bottom: Resulting brick poses.}
	\label{fig:align}
\end{figure}
To keep the wall structure intact during optimization, we add additional cost
terms for relationships between bricks that touch each other, which punish
deviations from their relative poses in the initialization:

{\footnotesize\begin{alignat}{4}
E_{i,j}^{R} &= \lambda_r\|R(\theta_i)\R{B}{b_i}(R(\theta_j)\R{B}{b_j})^{-1}\R{B}{b_j}\R{b_i}{B} -I\|_{F}^2, \\
E_{i,j}^{T} &= \lambda_t\|t(T(\theta_i,t_i)\T{B}{b_i}(T(\theta_j,t_j)\T{B}{b_j})^{-1}\T{B}{b_j} \T{b_i}{B})\|_2^2,
\end{alignat}}

where $||\cdot||_F$ denotes the matrix norm, and $\lambda_r, \lambda_t$ are
balancing factors. Note that these pairwise cost terms have equal strength
for all involved brick pairs.

As in the rough alignment phase, the parameters are optimized using Ceres
using the same termination criteria up to a maximum of 20 iterations.
The optimization takes around 0.15\,s on the onboard computer for one iteration with 20 bricks.
In addition, we compute a confidence parameter for each brick as the ratio of
found correspondences to expected visible points according to the rendered model.
\Cref{fig:align} shows an exemplary result of the entire pipeline.

\subsection{Experiments}

During the MBZIRC 2020 Finals, our UGV Bob performed in six arena runs.
We used the three rehearsal days to get familiar with the arena conditions,
fixed Wi-Fi issues, picked bricks in a semi-autonomous way and fine-tuned
our perception pipeline.
Unfortunately, in the first Challenge~2 competition run we had issues
regarding the gripper. We attempted over 15 times picking up an orange
brick with very promising perception results but were not successful.
We were unable to fix this problem since hardware changes were not allowed
during the competition run.
In the second competition run we were able to pick and store a green
brick successfully, but again scored zero points since our UGV was not able to drive accurately on the slope inside the arena to reach the wall position. Due to limited test time
we did not discover this problem earlier. Nevertheless, the points collected by our
UAV were enough to secure an overall second place.
In the final Grand Challenge, our UGV was assigned to first solve Challenge~3
(fire fighting) to maximize
the overall points of our team. After successfully solving Challenge~3, only two minutes
were left, which was not enough time to score any points in Challenge~2.

\begin{table}
\centering
\begin{threeparttable}
\caption{Build order optimization}
\label{tab:bob_strategy}
\setlength{\tabcolsep}{6pt}
\begin{tabular}{lrrrrrr}
\toprule
Method & \multicolumn{2}{c}{$|B|$} & \multicolumn{2}{c}{$d_{B}$ [m]} & \multicolumn{2}{c}{Runtime [s]} \\
\cmidrule (lr){2-3} \cmidrule (lr){4-5} \cmidrule (lr){6-7}
       & mean & stddev & mean & stddev & mean & stddev \\
\midrule
Optimal & 5.0 & 0.91 & 3.36 & 0.97 & 7.5 & 29.0 \\
Greedy & 5.5 & 1.10 & 5.63 & 1.86 & 0.0 & 0.0 \\
\bottomrule
\end{tabular}
Computing the build order ($B$) using our optimization versus a greedy approach
over 1000 randomly generated blueprints. The path length to reach all build positions is denoted as $d_B$.
\end{threeparttable}
\end{table}

After the competition, we evaluated two sub-systems of our UGV in our lab environment.
We compared our algorithm for optimizing the build order with a greedy strategy.
Using the greedy strategy, we take the best local solution, i.e. given a set of already built
bricks, we chose the next build position such that the number of bricks the UGV is able to build
is maximal. \cref{tab:bob_strategy} shows the results of both approaches on
1000 randomly generated blueprints. The optimization reduces the different build positions
needed from 5.5 to 5.0 on average and gives an even larger improvement regarding the
distance needed to be driven by 2.3\,m on average. Performing the optimization
takes on average 7.5\,s, which is feasible in our use case since it is done just once before the
competition run. Nevertheless, it is---as expected---much slower than the greedy approach
due to the exponential complexity.

\begin{figure}[b]
 \centering
﻿\resizebox{.95\linewidth}{!}{%
\begin{tikzpicture}
[font=\sffamily\Large]

\coordinate (r) at (18.5,0);
\coordinate (g) at (12.5,3);
\coordinate (b) at (0,3);
\coordinate (o) at (0,0);
\coordinate (l) at (20,3.7);

\draw[thick, black] (r) rectangle ($(r) + (3,2)$);
\draw[thin, red, rotate around={0.95:($(r) + (0.17,0.24)$)}] ($(r) + (0.17,0.24)$) rectangle ($(r) + (3.17,2.24)$);
\draw[thin, red, rotate around={1.72:($(r) + (0.11,0.17)$)}] ($(r) + (0.11,0.17)$) rectangle ($(r) + (3.11,2.17)$);
\draw[thin, red, rotate around={1.91:($(r) + (0.17,0.23)$)}] ($(r) + (0.17,0.23)$) rectangle ($(r) + (3.17,2.23)$);
\draw[thin, red, rotate around={0.38:($(r) + (0.16,0.11)$)}] ($(r) + (0.16,0.11)$) rectangle ($(r) + (3.16,2.11)$);
\draw[thin, red, rotate around={0:($(r) + (0.19,0.2)$)}] ($(r) + (0.19,0.2)$) rectangle ($(r) + (3.19,2.2)$);
\draw[thin, red, rotate around={-1.34:($(r) + (0.1,0.1)$)}] ($(r) + (0.1,0.1)$) rectangle ($(r) + (3.1,2.1)$);
\draw[thin, red, rotate around={-0.38:($(r) + (0.13,0.11)$)}] ($(r) + (0.13,0.11)$) rectangle ($(r) + (3.13,2.11)$);
\draw[thin, red, rotate around={2.29:($(r) + (0.19,0.2)$)}] ($(r) + (0.19,0.2)$) rectangle ($(r) + (3.19,2.2)$);
\draw[thin, red, rotate around={1.53:($(r) + (0.18,0.18)$)}] ($(r) + (0.18,0.18)$) rectangle ($(r) + (3.18,2.18)$);
\draw[thin, red, rotate around={2.1:($(r) + (0.17,0.19)$)}] ($(r) + (0.17,0.19)$) rectangle ($(r) + (3.17,2.19)$);

\draw[thick, black] (g) rectangle ($(g) + ((6,2)$);
\draw[thin, green, rotate around={-0.95:($(g) + (0.21,0.04)$)}] ($(g) + (0.21,0.04)$) rectangle ($(g) + (6.21,2.04)$);
\draw[thin, green, rotate around={2.01:($(g) + (0.14,0.16)$)}] ($(g) + (0.14,0.16)$) rectangle ($(g) + (6.14,2.16)$);
\draw[thin, green, rotate around={-1.05:($(g) + (0.2,-0.04)$)}] ($(g) + (0.2,-0.04)$) rectangle ($(g) + (6.2,1.96)$);
\draw[thin, green, rotate around={-0.95:($(g) + (0.2,0.04)$)}] ($(g) + (0.2,0.04)$) rectangle ($(g) + (6.2,2.04)$);
\draw[thin, green, rotate around={1.72:($(g) + (0.16,0.16)$)}] ($(g) + (0.16,0.16)$) rectangle ($(g) + (6.16,2.16)$);
\draw[thin, green, rotate around={0.76:($(g) + (0.13,0.05)$)}] ($(g) + (0.13,0.05)$) rectangle ($(g) + (6.13,2.05)$);
\draw[thin, green, rotate around={-2.67:($(g) + (0.23,-0.01)$)}] ($(g) + (0.23,-0.01)$) rectangle ($(g) + (6.23,1.99)$);
\draw[thin, green, rotate around={-1.05:($(g) + (0.2,-0.04)$)}] ($(g) + (0.2,-0.04)$) rectangle ($(g) + (6.2,1.96)$);
\draw[thin, green, rotate around={-0.38:($(g) + (0.19,0.1)$)}] ($(g) + (0.19,0.1)$) rectangle ($(g) + (6.19,2.1)$);
\draw[thin, green, rotate around={1.05:($(g) + (0.16,0.13)$)}] ($(g) + (0.16,0.13)$) rectangle ($(g) + (6.16,2.13)$);

\draw[thick, black] (b) rectangle ($(b) + (12,2)$);
\draw[thin, blue, rotate around={0.57:($(b) + (0.21,0.12)$)}] ($(b) + (0.21,0.12)$) rectangle ($(b) + (12.21,2.12)$);
\draw[thin, blue, rotate around={0.67:($(b) + (0.1,0.1)$)}] ($(b) + (0.1,0.1)$) rectangle ($(b) + (12.1,2.1)$);
\draw[thin, blue, rotate around={0.91:($(b) + (0.17,0.165)$)}] ($(b) + (0.17,0.165)$) rectangle ($(b) + (12.17,2.165)$);
\draw[thin, blue, rotate around={0.93:($(b) + (0.17,0.17)$)}] ($(b) + (0.17,0.17)$) rectangle ($(b) + (12.17,2.17)$);
\draw[thin, blue, rotate around={1.1:($(b) + (0.26,0.18)$)}] ($(b) + (0.26,0.18)$) rectangle ($(b) + (12.26,2.18)$);
\draw[thin, blue, rotate around={0.91:($(b) + (0.13,0.17)$)}] ($(b) + (0.13,0.17)$) rectangle ($(b) + (12.13,2.17)$);
\draw[thin, blue, rotate around={0.91:($(b) + (0.22,0.15)$)}] ($(b) + (0.22,0.15)$) rectangle ($(b) + (12.22,2.15)$);
\draw[thin, blue, rotate around={1.38:($(b) + (0.16,0.16)$)}] ($(b) + (0.16,0.16)$) rectangle ($(b) + (12.16,2.16)$);
\draw[thin, blue, rotate around={0.05:($(b) + (0.165,0.06)$)}] ($(b) + (0.165,0.06)$) rectangle ($(b) + (12.165,2.06)$);
\draw[thin, blue, rotate around={0.19:($(b) + (0.22,0.05)$)}] ($(b) + (0.22,0.05)$) rectangle ($(b) + (12.22,2.05)$);

\draw[thick, black] (o) rectangle ($(o) + (18,2)$);
\draw[thin, orange, rotate around={-0.38:($(o) + (0.11,-0.07)$)}] ($(o) + (0.11,-0.07)$) rectangle ($(o) + (18.11,1.93)$);
\draw[thin, orange, rotate around={-0.32:($(o) + (0.16,-0.07)$)}] ($(o) + (0.16,-0.07)$) rectangle ($(o) + (18.16,1.93)$);
\draw[thin, orange, rotate around={-0.35:($(o) + (0.15,-0.04)$)}] ($(o) + (0.15,-0.04)$) rectangle ($(o) + (18.15,1.96)$);
\draw[thin, orange, rotate around={0.45:($(o) + (0.12,0.14)$)}] ($(o) + (0.12,0.14)$) rectangle ($(o) + (18.12,2.14)$);
\draw[thin, orange, rotate around={0.73:($(o) + (0.05,0.17)$)}] ($(o) + (0.05,0.17)$) rectangle ($(o) + (18.05,2.17)$);
\draw[thin, orange, rotate around={0.22:($(o) + (0.16,0.02)$)}] ($(o) + (0.16,0.02)$) rectangle ($(o) + (18.16,2.02)$);
\draw[thin, orange, rotate around={-0.92:($(o) + (0.22,-0.09)$)}] ($(o) + (0.22,-0.09)$) rectangle ($(o) + (18.22,1.91)$);
\draw[thin, orange, rotate around={0.54:($(o) + (0.05,0.17)$)}] ($(o) + (0.05,0.17)$) rectangle ($(o) + (18.05,2.17)$);
\draw[thin, orange, rotate around={0.13:($(o) + (0.14,0.07)$)}] ($(o) + (0.14,0.07)$) rectangle ($(o) + (18.14,2.07)$);
\draw[thin, orange, rotate around={0.45:($(o) + (0.16,0.14)$)}] ($(o) + (0.16,0.14)$) rectangle ($(o) + (18.16,2.14)$);

\node (x) at ($(l) + (1.1,0)$){x} ;
\node (y) at ($(l) + (0,1.1)$){y} ;
\draw[->, line width=0.3mm](l)  -- ($(l) + (0,0.8)$);
\draw[->, line width=0.3mm](l)  -- ($(l) + (0.8,0)$);

\end{tikzpicture}
}
\vspace*{-2ex}
 \caption{UGV picking robustness. Ground truth place pose (black) and ten test results for each brick type.}
 \label{fig:bob_eval_grasping}
\end{figure}
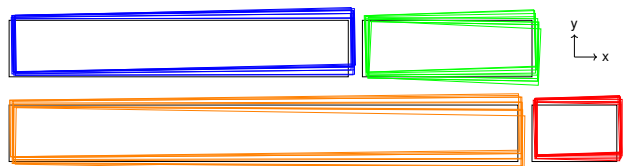

In a second lab experiment, we evaluated the precision and repeatability of
picking up bricks from a pile (see \cref{fig:bob_eval_grasping}). We placed four bricks in front of the robot
similar to the competition setup. Each test consists
of scanning the piles, picking the brick with the highest confidence, and placing it
at a predefined pose. We calculated the mean translation and rotation error
compared to a perfectly aligned center-grasped brick. Only a rough estimation of the pile
location was provided to the system. We repeated the test ten times while changing the brick
horizontal positions by up to 5\,cm and the rotation by up to 10$^\circ$ around the vertical axis.
\Cref{tab:bob_grasping} shows the mean results per brick.
The resulting mean error could be further decreased by investing more time calibrating
the whole system; nevertheless, it is sufficient to place the bricks reliably
into the storage system. The very low standard deviation in both rotation and translation
shows that our perception and grasping pipeline has a high repeatability and is very
robust.

\begin{table}
\begin{threeparttable}
 \caption{End-To-End Brick Manipulation Precision}
\label{tab:bob_grasping}
\centering
\setlength{\tabcolsep}{6pt}
\begin{tabular}{lrrrrrr}
\toprule
Brick & \multicolumn{2}{c}{Translation x [cm]} & \multicolumn{2}{c}{Translation y [cm]} & \multicolumn{2}{c}{Yaw [$^\circ$]} \\
\cmidrule (lr) {2-3} \cmidrule (lr){4-5} \cmidrule (lr){6-7}
      & mean & stddev & mean & stddev & mean & stddev \\
\midrule
Red    & 1.52 & 0.36 & 1.49 & 0.37 & 1.26 & 0.80 \\
Green  & 1.94 & 0.33 & 0.67 & 0.43 & 1.26 & 0.68 \\
Blue   & 1.82 & 0.49 & 0.53 & 0.21 & 0.76 & 0.40 \\
Orange & 1.34 & 0.65 & 0.36 & 0.39 & 0.45 & 0.24 \\
\bottomrule
\end{tabular}
Placement error from perceiving, picking, and placing. Ten tests per color.
\end{threeparttable}
\end{table}

\FloatBarrier
\section{UAV Solution}

Since the initial rules specified a shared wall where UAVs and UGVs could
collaborate, we concentrated our efforts on the UGV design.
In a late rule revision UGV and UAV walls were separated, making it clear to us
that UAV points had to be scored in order to win.
Our UAV design thus focused on a minimal solution that could achieve
almost full points: We decided to ignore the orange bricks of 1.8\,m length,
which were intended to be carried by two UAVs. Our system should support the
red (0.3\,m), green (0.6\,m), and blue (1.2\,m) bricks.

\subsection{Hardware Design}

Because of the weight of the larger bricks and their size, we decided to
use a large UAV, the DJI Matrice 600 (M600), for this task. The M600 offers
sufficient payload and battery life (roughly 20\,min in our configuration).

A key component for aerial manipulation is the robotic gripper. UAVs pose unique
constraints when compared with ground-based manipulation.
The gripper has to be light-weight in order to fit inside the payload constraints.
Furthermore, a certain flexibility and mechanical compliance is desired for two reasons:
First, this allows a grasp to succeed even if the approach was not fully precise.
Secondly, a rigid connection between the UAV and the ground can be
very dangerous, since UAVs usually tightly and very dynamically control their attitude in order to hold position.
One can easily imagine situations where the UAV has to drastically change attitude in
response to wind gusts and of course hindrance by the gripper system should be
limited.
However, during the placement phase of the pick-and-place operation, we require very
precise control of the target object. Here---at least while the target object is still
in the air---we want a rigid attachment to the UAV. To resolve these seemingly
contradicting goals, we designed the gripper system to be rigid only while load is applied,
i.e. the brick is hanging below the UAV.

\begin{figure}[b]
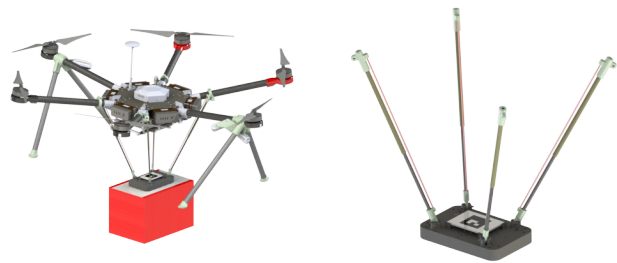

 \centering %
 \definecolor{gripperbg}{RGB}{102,171,184}
 \includegraphics[height=3.5cm,clip,trim=0 20 0 100]{images/lofty/cad_lofty2.png}\hfill%
 \includegraphics[height=3.5cm,clip,trim=50 70 20 20]{images/lofty/cad_gripper.png}%
 \vspace{-2ex}
 \caption{UAV hardware design.
  Left: Full assembly.
  Right: Magnetic gripper with passive compliance. The four telescopic
  rods are shown in fully extended configuration.}
 \label{fig:lofty_cad}
\end{figure}

Our gripper design (see \cref{fig:lofty_cad}) consists of four carbon fiber telescopic rods, which hold a
plate equipped with 8 electromagnets (similar to the UGV gripper) below the UAV. When the rods are fully extended, the gripper
plate is in a fixed pose and can only move upwards. The more the gripper plate
is pressed upwards (e.g. due to contact with a brick), the more it can move sideways
and rotate due to the gained movement range in each rod.
The gripper is equipped with a switch to detect successful grasping.

Since the standard foldable landing legs on the M600 would interfere with the
gripper, we replaced them with fixed landing legs (see \cref{fig:lofty_cad}).

\subsection{Brick and Wall Perception}

The competition task involves two perception challenges: finding and precisely
localizing the bricks and localizing with respect to the target wall.
Similarly to the gripper system, the UAV places unique constraints on the
perception system. Because the gripper is mounted directly beneath the UAV,
any observation of a brick close to the gripper must be done from the side.
The necessary off-center mounting of the sensor severely limits the sensor
weight. A 3D-LiDAR as used in the UGV is too heavy.
We chose the Intel RealSense D435 RGB-D
camera as a primary sensor for its light weight and its capability to work in sunlight.
To achieve good coverage of the terrain below
the UAV and to be able to observe large parts of the wall during the placement
process, we mounted three D435 sensors on the UAV (see \cref{fig:lofty_cad,fig:lofty_perception}).
In contrast to the UGV solution, the UAV solution needs to be real-time capable
to allow tracking during approach.

\subsubsection{Brick Detection \& Pose Estimation}

\begin{figure*}[t]
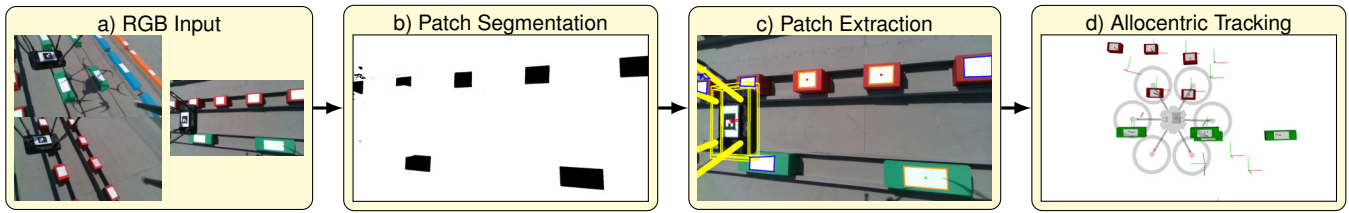

 \centering
 \begin{tikzpicture}[
    font=\footnotesize\sffamily\footnotesize,
    img/.style={draw=black, rounded corners, align=center, fill=yellow!20}
  ]
  \node[img] (input) {a) RGB Input\\
  \parbox{.11\linewidth}{%
	\includegraphics[width=\linewidth]{images/lofty/brick/scene1/cam1.png}
	\includegraphics[width=\linewidth]{images/lofty/brick/scene1/cam2.png}
  }\hspace{.11cm}%
  \parbox{.1\linewidth}{%
    \includegraphics[width=\linewidth]{images/lofty/brick/scene1/input.png}
  }
  };

  \node[img,right=0.4cm of input] (segmentation) {b) Patch Segmentation\\\includegraphics[width=0.22\linewidth,frame]{images/lofty/brick/scene1/segmentation.png}};

  \node[img,right=0.4cm of segmentation] (vis) {c) Patch Extraction\\\includegraphics[width=0.22\linewidth]{images/lofty/brick/scene1/vis.png}};

  \node[img,right=0.4cm of vis] (rviz) {d) Allocentric Tracking\\\includegraphics[width=0.22\linewidth,frame]{images/lofty/brick/scene1/rviz_crop.png}};

  \draw[-latex,very thick] (input) -- (segmentation);
  \draw[-latex,very thick] (segmentation) -- (vis);
  \draw[-latex,very thick] (vis) -- (rviz);
 \end{tikzpicture}
 \vspace{-1.5em}
 \caption{Camera-based UAV brick perception pipeline.
   a)~Input frames from all cameras.
   b)~White patch segmentation.
   c)~Patch corner extraction \& pose estimation.
     Patch contours in orange (verified) and blue (wrong shape).
     Brick type is indicated by a colored center point.
     The gripper is overlaid in yellow.
   d)~Tracking of detections from all three cameras in GPS frame. Detections
     are shown as bricks, while tracked hypotheses are shown as coordinate axes.
 }
 \label{fig:lofty_perception}
 \vspace{-.5cm}
\end{figure*}

The gripper is visible in all camera images and would lead to confusion with
bricks. For this reason, we mounted an ArUco marker~\citep{romero2018speeded}
on it. The marker pose can be efficiently estimated in each of the three
cameras and is low-pass filtered to obtain a robust estimate of the gripper
pose below the UAV. Pixels in the immediate vicinity to the detected gripper are
discarded for the following processing steps.

Since the white patches on the bricks are quite distinctive (see \cref{fig:lofty_perception}),
we use them to detect the bricks and estimate their pose. In a first step,
we convert the input image (resolution 1280$\times$720) to the HSV color space.
To detect high-saturation pixels (the colored bricks) in the neighborhood,
we downsample the input image to half resolution and run a box filter with
kernel size 290$\times$290 to obtain a local saturation average $\bar{S}$
and local value average $\bar{V}$.
A pixel $p$ is classified as \textit{patch}, if $S(p) < \bar{S}(p) - \lambda_S  \wedge  V(p) > \bar{V}(p) + \lambda_V$,
or, in other words, the saturation is less than the local average and the value
(brightness) is larger than the local average, by user-specified thresholds.
This simple segmentation method is modeled after the ones used for detecting
chessboard patterns and leads to highly robust performance (see \cref{fig:lofty_perception}).

Contours with exactly four corners (after contour simplification) are processed
further: We check that each corner has a \textit{patch} pixel on the inside
and a high-saturation pixel on the outside at a specified distance $d=4$ pixels.
The high-saturation pixels on the outside are independently classified into the
four possible colors. If all agree, the brick is detected.

Finally, a PnP solver is used to determine the 6D pose of the brick from
the recovered 2D-3D correspondences. Here, we assume that the longer
side in the 2D image corresponds to the longer brick side in 3D---an assumption
which is only violated at extreme viewing angles.
To fuse the detections from all three cameras and to track bricks over time,
we apply a basic Multi-Hypothesis Tracking (MHT) method with one Kalman filter
per hypothesis.

\subsubsection{Wall Localization}

\begin{figure}[t]
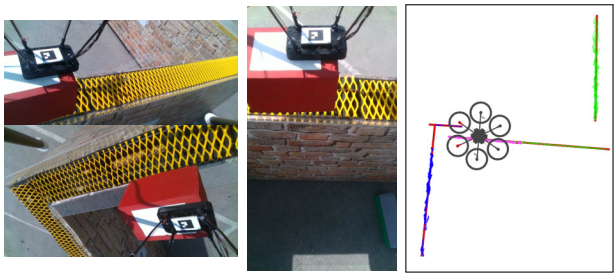

 \centering
 \parbox{3.1cm}{%
 \includegraphics[width=\linewidth]{images/lofty/wall/scene2_cam2.png}\vspace{-1em}
 \includegraphics[angle=180,width=\linewidth]{images/lofty/wall/scene2_cam1.png}}
 \parbox{1.98cm}{%
 \includegraphics[width=\linewidth]{images/lofty/wall/scene2_brio_rot.png}
 }\hspace{0em}
 \parbox{2.74cm}{%
 \includegraphics[angle=180,width=\linewidth,clip,frame,trim=220 80 270 80]{images/lofty/wall/scene2_rviz.png}
 }
 \caption{Wall localization. Left: RGB images from all three cameras.
 Right: Top-down view of detected wall points per camera (green, blue, purple)
 and detected wall segments (red lines).}
 \label{fig:wall_perception}
\end{figure}

After grasping a brick, the UAV needs to bring it to the wall and place it.
Before the competition in Abu Dhabi very little was known about the wall
except its geometric shape: Four segments of four meters length and 1.7\,m height,
arranged in a ``W'' shape.
Especially the top part, which is easily visible from the UAVs perspective,
was highly problematic: No information about visual appearance was available,
visibility of the top covering (gridding with unspecified mesh size) in our
depth sensors was unknown, and later on it would be covered with placed
or dropped bricks.
We decided to focus on the side walls instead, which where specified as more
or less flat surfaces. Especially the side-facing cameras would be able to see
the side walls during close approach.

Consequently, our wall perception module estimates the height above ground
from the depth image of the downward-facing camera. Points from each camera
are then filtered so that only points above 1.0\,m and below 1.7\,m remain.
The data is then projected to 2D, where lines can be extracted using RANSAC.
Each line, if fit correctly, corresponds to a side view of one wall segment (see \cref{fig:wall_perception}).

The system is initialized with a user-specified
initial wall pose, which serves as the search pose.
Any time two parallel line segments of valid length with 4\,m distance are found, the wall pose is
updated. Under the assumption that the wall did not rotate 180$^\circ$, this is unambiguous.

During close approach, the UAV targets a specific place pose on one of the wall
segments. The detected segment closest to the expected segment pose is
identified and the goal position is projected onto this segment.

\subsection{High-level Control}

Similar to the UGV, the high-level control module is implemented in a FSM framework. It is supplied with the target wall pattern as defined
by the organizers of the competition.
The basic cycle of events is designed as follows:
\begin{enumerate}
 \item Fly to the last known pile pose and fly a search pattern until the next
   brick requested by the pattern is found.
 \item Grasp the brick and lift it.
 \item Fly to the target position (relative to the last known wall pose) and
   look for a wall segment.
 \item Approach the projected position on the wall segment and place
   the brick.
\end{enumerate}

Similarly to our MBZIRC 2017 approach~\citep{beul2019team},
we utilize a ``cone of descent'' during grasping and placement, in
which the UAV is allowed to descend towards the target pose.
If it drifts outside
of the cone, it has to stay at that height until the disturbance has been rejected.
The cone angle is $10^\circ$ with a hysteresis
of $3^\circ$ to prevent oscillations. The cone was shifted such that
at the target height it had a radius of 9\,cm, which was determined as
the maximum deviation that would still allow successful magnetic grasping.

\subsection{Experiments}

During the MBZIRC 2020 Finals, our UAV Lofty performed in six arena runs:
three rehearsal runs, two Challenge~2 runs, and the final Grand Challenge run.

We used the rehearsal runs to get used to the conditions in Abu Dhabi and
continuously improved our pick success rate.
During our first Challenge~2 run, we only picked one red and one green brick
due to difficulties with our magnetic gripper. Both bricks were dropped
close to, but not on the wall due to wall tracking problems. The wall
tracking module had not been tested until this point due to short development time
and lack of suitable testing opportunities at the competition.
After improving our gripper overnight, we managed to pick four red bricks
and one green brick and placed two red bricks successfully during our second Challenge~2 run. The other bricks were sadly dropped right next to the wall due to
another wall tracking problem.
This run was scored as 1.33~points, which secured a second place in Challenge~2,
next only to the Prague-Pennsylvania team.

In the Grand Challenge, Lofty managed to pick a red brick, but placed it
a bit too high and it fell off the wall. After a longer pause to allow our
Challenge~1 UAV to operate, it started again and picked up a green brick.
Sadly, it falsely detected a W-shaped wall behind the arena netting.
Due to a rushed setup sequence, the geofencing was not configured correctly and
did not prevent Lofty from flying into the net.
After a short unsuccessful rescue attempt during a reset, we had to leave it
there for the rest of the Grand Challenge.

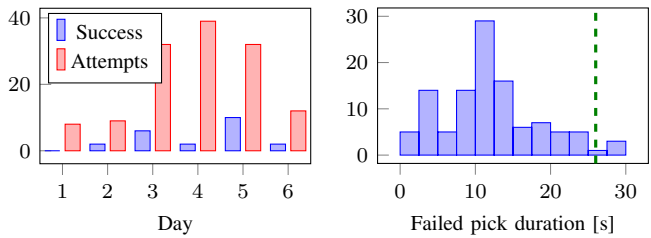
\begin{figure}[t]
 \centering
 \vspace{1ex}
 \begin{tikzpicture}[font=\footnotesize]
  \begin{axis}[ybar, bar width=0.2cm, legend pos=north west, xlabel=Day, width=.6\linewidth, height=3.65cm, xtick distance=1,tick pos=left]
   \addplot+ table [x=day,y=success] {\dailysuccess};
   \addlegendentry{Success}

   \addplot+ table [x=day,y=total] {\dailysuccess};
   \addlegendentry{Attempts}
  \end{axis}
 \end{tikzpicture}\hfill
 \begin{tikzpicture}[font=\footnotesize]
  \begin{axis}[ybar, width=.6\linewidth, height=3.65cm, xlabel={Failed pick duration [s]},tick pos=left]
   \addplot +[
    hist={
        bins=12,
        data min=0.0,
        data max=30.0
    }
   ] table [y index=0] {data/lofty_pick_fail_times.txt};
   \draw[green!50!black,very thick,dashed] (axis cs:26, \pgfkeysvalueof{/pgfplots/ymin})
    --
    (axis cs:26, \pgfkeysvalueof{/pgfplots/ymax});
  \end{axis}

 \end{tikzpicture}

 \vspace{-1em}
 \caption{Picking robustness. Left: Success rate over the duration of the competition.
 Right: Histogram of failed pick durations. The average successful pick duration
 is shown in green.}
 \label{fig:pick_success_rate}
 \vspace{-1em}
\end{figure}

Overall, Lofty executed 132 pick attempts in Abu Dhabi, of which 22 were
successful, which gives a success rate of 16.7\%.
Since a failed attempt took 12\,s on average, this limited the number of attempts
we had for placing bricks on the wall.
The number of pick attempts increased over the duration of the competition
(see \cref{fig:pick_success_rate}) as the rest of the system became more robust.
There are two peaks in the duration histogram for failed picks: One at roughly
three seconds which corresponds to tracking failures during the initial approach,
and a larger one around 10\,s, which corresponds to misaligned picks or magnet
failures.

\pgfplotstableread{
day success total percentage
1   0       8     0.0
2   2       9     22.2
3   6       32    18.8
4   2       39    5.13
5   10      32    31.3
6   2       12    16.7
}\dailysuccess

\section{Conclusion}

We take the opportunity to identify key strengths and weaknesses of our system
and development approach. We also want to identify aspects of the competition
that could be improved to increase scientific usefulness in the future.

First of all, this edition of the MBZIRC suffered from low team performance,
to the extent that the Grand Challenge price money was not paid out on recommendation of the
jury. This underperformance of all teams points to systematic
issues with the competition.
From the perspective of participants, we think the late changes of the rules have certainly contributed to this situation.
A pre-competition event such as the Testbed in the DARPA Robotics Challenge
can help to identify key issues with rules and material early in the competition
timeline.
Another issue was the required effort to participate in all the different
sub-challenges. MBZIRC 2020 defined seven different tasks---ideally,
one would develop specialized solutions for all of these.
Focusing the competition more on general usability, i.e. defining multiple tasks that
can and should be completed by one platform, would lower the barrier for participants.

Regarding our system, we saw very little problems with our hardware design---both
robots could have scored their theoretical maximum.
After solving initial problems with our magnets,
especially the passive UAV gripper turned out to be an advantage over other teams,
who could not manipulate the heavier bricks. The UGV brick perception provided
reliable brick poses during the competition and during lab experiments.

The biggest issue shortly before and during the competition was unavailable
testing time. Robust solutions require full-stack testing under competition
constraints. Since we postponed many design decisions until the rules were settled,
our complex design could not be tested fully. In the end, simpler designs with fewer components, which would have
required less thorough testing, could have been more successful in the short available time frame.

We presented a UGV-UAV system for autonomous wall building, which
successfully competed at the MBZIRC 2020. We will continue research into
aerial and terrestrial manipulation and further UAV-UGV cooperation.

\printbibliography

\end{document}